\documentclass[conference]{IEEEtran}

\usepackage{amssymb,amsmath}
\usepackage{cite}
\usepackage{subcaption}
\usepackage{graphicx}
\usepackage{psfrag}
\usepackage{url}
\usepackage[latin1]{inputenc}
\usepackage[absolute,overlay]{textpos}
\usepackage{gensymb}
\usepackage{cases}
\usepackage{bm}
\usepackage{graphicx}
\usepackage[font=footnotesize]{caption}
\usepackage[font=footnotesize]{subcaption}
\usepackage{float}
\usepackage[linesnumbered,ruled,lined]{algorithm2e}

\usepackage{tabularx}
\usepackage{array}

\allowdisplaybreaks
\usepackage{tikz}
\usepackage{booktabs,multirow}
\usetikzlibrary{patterns,arrows}
\definecolor{color1}{RGB}{237, 191, 193}
\definecolor{color2}{RGB}{229, 153, 157}
\definecolor{color3}{RGB}{225, 123, 116}
\definecolor{color4}{RGB}{10,10,200}
\definecolor{color5}{RGB}{203,52,38}

\usepackage{pgf}

\usepackage[nocomma]{optidef}

\usepackage{algorithmic}

\usepackage{multicol}
\usepackage{amsthm}

\usepackage{booktabs}
\usepackage{color,soul}

\begin{document}

\title{{\it Fed-Sophia}: A Communication-Efficient Second-Order Federated Learning Algorithm

\thanks{Identify applicable funding agency here. If none, delete this.}}

\author{
	\IEEEauthorblockN{Ahmed Elbakary$^{1,2}$, Chaouki Ben Issaid$^1$, Mohammad Shehab$^1$, Karim Seddik$^2$, Tamer ElBatt$^{3,4}$, Mehdi Bennis$^1$ \\
	}
	\IEEEauthorblockA{$^1$Centre for Wireless Communications (CWC), University of Oulu, Finland\\
	$^2$Department of Electronics and Communications Engineering, American University in Cairo, Egypt\\
        $^3$Department of Computer Science and Engineering, American University in Cairo, Egypt\\ 
        $^4$Department of Electronics and Communications Engineering, Cairo University, Egypt 
    }
    }

%Emails: \{ahmed.elbakary, chaouki.benissaid, mohammad.shehab, mehdi.bennis\}@oulu.fi}, kseddik@aucegypt.edu,

\maketitle

\begin{abstract}
Federated learning is a machine learning approach where multiple devices collaboratively learn with the help of a parameter server by sharing only their local updates. While gradient-based optimization techniques are widely adopted in this domain, the curvature information that second-order methods exhibit is crucial to guide and speed up the convergence. This paper introduces a scalable second-order method, allowing the adoption of curvature information in federated large models. Our method, coined Fed-Sophia, combines a weighted moving average of the gradient with a clipping operation to find the descent direction. In addition to that, a lightweight estimation of the Hessian's diagonal is used to incorporate the curvature information. Numerical evaluation shows the superiority, robustness, and scalability of the proposed Fed-Sophia scheme compared to first and second-order baselines.
\end{abstract}


\section{Introduction}

Recent advances in edge computing have paved the way to enable new machine learning paradigms (ML). One promising paradigm is federated learning (FL) \cite{mcmahan2017communication},\cite{kairouz2021advances}, where devices preserve their local data and only share, with a parameter server (PS), their local models. Privacy preservation and data locality are two major benefits of FL, in contrast to traditional centralized ML models, which require the entire dataset to be stored in a central cloud. On the contrary, FL relies on a two-stage approach. First, each device performs local training independently, using its own local data. After one or more iterations of local training, each device sends its local model or gradient to the PS. Afterward, the models are aggregated using one of the known aggregation schemes in the literature \cite{qi2023model}. Finally, after aggregation, the PS shares the refined model with all devices for enhanced performance. This process continues in an iterative manner over multiple communication rounds until convergence. The ultimate goal of FL is to reach a consensus and find a global model that solves the task at hand. 

When using first-order methods, clients employ the gradient to find a descent direction that could guide the convergence. Meanwhile, the default second-order optimization technique is Newton's method, which requires computing the full Hessian and using it as a pre-conditioner for the gradient. In our work, instead of computing the full Hessian as in Newton's method, which comes with a heavy computational cost, we rely on an estimation of the Hessian's diagonal to incorporate the curvature information across each dimension into the convergence process. One problem stemming from Newton's method is that it might converge to a saddle point instead of a local minimum. Another issue with Hessian-based methods is that the Hessian entries might mislead the convergence process by capturing non-accurate curvature information due to sudden changes in the loss function's curvature\cite{boyd2004convex}. To mitigate these problems, a clipping operation is introduced to guard against non-positive entries of the Hessian. 

% Our work is mainly inspired and builds upon \cite{liu2023sophia}, where the direction update is the gradient's moving average divided by the estimated Hessian's moving average. 

The main contributions of this paper can be summarized as follows
\begin{itemize}
  \item We propose Federated Sophia (Fed-Sophia), a scalable second-order method that enables large models to make use of the curvature information of the loss function. In addition to that, Fed-Sophia can handle heterogeneous settings where the data distribution differs among devices. 
  \item The proposed method uses a lightweight estimation of the Hessian and only shares the model's parameter vector, making it computation and communication-efficient. 
  \item Several numerical experiments are conducted to show that Fed-Sophia outperforms other baselines for solving an image classification task using convolutional neural network (CNN) or multi-layer perception (MLP) models. Furthermore, we compare the energy consumption and the carbon footprint of Fed-Sophia to other baselines. 
\end{itemize}
The rest of this paper is organized as follows. We highlight some of the available methods in Section \ref{REW}. In Section \ref{System}, the problem formulation and the system model are introduced. We introduce our proposed method in Section \ref{METHOD}. The details of our experimental results are laid out in Section \ref{RES}, where the proposed method is compared to other baselines. %on an image classification task.

\section{Related work} \label{REW}

In what follows, we highlight some differences between first and second-order methods and motivate the need for scalable optimizers in the FL setting.
\subsection{First-order FL Methods}
Federated averaging (FedAvg) introduced a generalization of stochastic gradient descent (SGD). The underlying idea is to have every device run several local updates before sending the local model parameters to the PS. The server then averages the local models' parameters and sends back the refined, global model parameters. In \cite{reddi2020adaptive}, the authors proposed a federated version of a set of popular centralized optimization methods like AdaGrad\cite{duchi2011adaptive} and Adam\cite{kingma2014adam}, using the same two-stage approach employed in FedAvg. To handle data heterogeneity, Scaffold \cite{karimireddy2020scaffold} used variance reduction to overcome the drift in local updates at each device. In \cite{haddadpour2021federated}, the authors reduced the communication overhead and the data heterogeneity using both compression and gradient tracking. The fundamental problem with first-order methods is their inability to capture the curvature information of the loss landscape, which leads to a uniform step size across all dimensions of the loss function.

% \cite{li2020federated} proposed FedProx to solve the same issue of data heterogeneity. 

\subsection{Second-order FL Methods}
Second-order methods have the advantage of capturing the effect of the curvature information of the loss function in the training process. FedNL \cite{safaryan2022fednl} is the first framework for using Newton's method in FL settings. The paper introduced compression techniques to improve communication efficiency between the PS and the devices. Still, the problem of computing the full Hessian at each device persists, making the idea impractical for large models. FedNew \cite{elgabli2022fednew} added another layer of privacy by hiding the gradient information. It also proposed stochastic quantization as a solution for the problem of communicating the Hessian with the PS while still approximating the inverse-Hessian-gradient product. FLECS\cite{agafonov2022flecs} proposed a framework for second-order methods, utilizing a lower dimensional representation of the Hessian to make the computation lightweight. A major bottleneck in this idea is that the server has to save the full Hessian. Meanwhile, DONE \cite{dinh2022done} uses an approximate Newton-type method and employs Richardson's iteration to find an approximation for the search direction at each client. Furthermore, DONE requires a large number of local iterations to converge, making it unreliable for practical settings.

\subsection{Optimizers in Deep Learning}
The choice of an optimizer in deep learning is often a task-dependent case. For example, SGD has dominated the computer vision field, while Adam\cite{kingma2014adam} excels in transformers-based language models. On the other hand, second-order methods are not widely adopted in training deep neural networks (DNN). One attempt to apply second-order methods is AdaHessian\cite{yao2021adaHessian}, a second-order method that estimates the Hessian's diagonal using Hutchinson's method\cite{hutchinson1989stochastic} with a spatial averaging mechanism to control misleading Hessian entries. In \cite{liu2023sophia}, the authors proposed Sophia, a second-order optimizer that exploits a lightweight estimation for the diagonal of the Hessian to incorporate curvature information into the optimization process. Originally introduced as an optimizer for large language models (LLMs), one advantage of Sophia is that it does not estimate the diagonal of the Hessian in each iteration, a key feature that can be well leveraged in FL settings.

% \subsection{Contributions}

\section{System Model and Problem Formulation}\label{System}
Consider a network of $N$ devices and a PS, where each device can communicate with the PS only. Each device has its local data set $\mathcal{D}_i(\bm{X}, \bm{y})$ and loss function $f_i(\bm{\theta}): \mathbb{R}^d \xrightarrow{} \mathbb{R}$, where $\bm{X}$ is the features matrix, $\bm{y}$ is the label vector, and $\bm{\theta}$ is the ML model we aim to train, e.g., the weights and biases of a NN. Devices train a local model $\bm{\theta}_i \in \mathbb{R}^d$ and send the trained model to the PS. The server can aggregate the information sent from the devices using some aggregation scheme to guide the convergence process. After aggregation, the server sends back the updated information to all clients, which starts another round of local training. This process continues until convergence. The objective of the training process is to perform empirical risk minimization in the form
\begin{align}
\min_{\bm{\theta} \in \mathbb{R}^d}{f(\bm{\theta}) \triangleq \frac{1}{N} \sum_{i=1}^N f_i(\bm{\theta})}.  
\end{align}

%where $h_i(x) = \frac{1}{m} \sum\limits_{j=1}^m l(x, (a_j, b_j)) + \frac{\lambda}{2} ||x||^2$.\newline

% \textbf{Assumption 1.} The function $f: \mathbb{R^d} \xrightarrow{} \mathbb{R} is twice continuously differentiable$

% The above assumption is required to use in any second-order information that needs to compute the Hessian or approximate it using any techniques such as the diagonal Hessian estimation.\newline

In the remainder of this paper, let $\bm{H}_i \triangleq \nabla^2 f_i(\bm{\theta})$ and $\bm{g}_i \triangleq \nabla f_i(\bm{\theta})$ denote the Hessian and the gradient of the loss function of the $i$th device, respectively. The entries of the Hessian matrix could be divided into two classes: the diagonal elements and the off-diagonal elements. The diagonal elements indicate the curvature information across each dimension. High values for diagonal elements mean steep curvature, which requires smaller steps. Meanwhile, small values reflect a flat dimension, in which case faster steps could be adopted. On the other hand, off-diagonal elements represent the cross-curvature information. To illustrate the effect of the Hessian's elements, let's consider the following loss function
\begin{align}
f(\theta_1, \theta_2) = \theta_1^2 + 2\theta_1\theta_2 + 3\theta_2^2,
\end{align}
where $\bm{\theta} = [\theta_1, \theta_2]$ is the model we aim to learn. The Hessian matrix for this function is given by   
\begin{align}
    \bm{H} = \nabla^{2} f(\bm{\theta}) = \begin{bmatrix}
\frac{\partial^2 f}{\partial \theta_1^2} & \frac{\partial^2 f}{\partial \theta_1 \partial \theta_2} \\
\frac{\partial^2 f}{\partial \theta_2 \partial \theta_1} & \frac{\partial^2 f}{\partial \theta_2^2}
\end{bmatrix}
= \begin{bmatrix}
2 & 2 \\
2 & 6
\end{bmatrix}.
\end{align}

\begin{figure}
    \centering
    \includegraphics{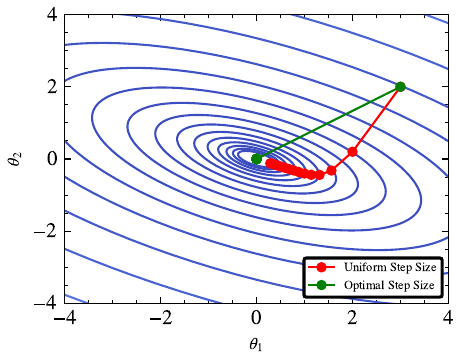}
    \caption{The effect of the Hessian-based step size against the gradient-based method.}
    \label{fig:step_size}
\end{figure}

The optimal step size to take across a dimension should be proportional to the Hessian entries corresponding to those dimensions. For example, the optimal step size for $\theta_1$ and $\theta_2$ should be proportional to $1/2$ and $1/6$, respectively. Those entries are the diagonal of the Hessian. Figure \ref{fig:step_size} demonstrates the effect of the optimal step size on the convergence. As we can see, the gradient-based method using uniform step size takes a lot of steps to converge, while a Hessian-based method can converge in two steps.

\section{Proposed Algorithm}\label{METHOD}
Our proposed algorithm, Fed-Sophia, is a second-order method that leverages the Sophia optimizer \cite{liu2023sophia} into the FL setting. The algorithm's computational and communication efficiency is attributed to two primary factors: (i) the lightweight estimation of the Hessian matrix, which, owing to its diagonal structure, is analogous to computing a vector, and (ii) the Hessian estimation does not occur in every communication round. Similar to FedAvg, every device in Fed-Sophia shares its model's parameters with the PS. Then, the server aggregates the parameters using their average as follows
\begin{align}
\bm{\Theta}^{k+1} = \frac{1}{N}\sum_{i=1}^{N}\bm{\theta}_i^{k},   
\end{align}
where $\bm{\Theta}$ is the server's parameters at iteration $(k+1)$, and $\bm{\theta}_i$ is the model's parameters of the $i$th device. As shown in\cite{mcmahan2017communication,lee2023partial}, this idea of averaging the models works surprisingly well with over-parameterized NNs.

\begin{algorithm}[t]
    \caption{Federated Sophia (Fed-Sohpia)}
    \label{alg:federated_hess_opt}
    \begin{algorithmic}[1]
        \STATE {\bf Parameters:} learning rate $\eta$, number of local iterations $J$, hyperparameters $(\lambda, \beta_1, \beta_2, \epsilon, \rho, \tau)$.
        \STATE {\bf Initialize:} $\bm{\Theta}^0$, $\bm{m}^0 = \bm{0}$, $\bm{h}_i^{-\tau} = \bm{0}$
        \FOR{communication round $k = 0$ to $K$}
            \FOR{each device $i \in [N]$}
                \STATE Receive the global model $\bm{\Theta}^k$ and set the local\newline model as $\bm{\theta}_i^k = \bm{\Theta}^k$ \\
                % \textcolor{red}{Do not forget to add the local iterations in the pseudo-code.}
                \FOR{local iterations $j = 0$ to $J$}
                \STATE Compute the local gradient $\bm{g}_{i}^k = \nabla h_{i}(\bm{\theta}_i^k)$
                \STATE Update the local moving average of the\\ gradient:  $\bm{m}_{i}^k = \beta_1 \bm{m}_{i}^{k-1} + (1 - \beta_1) \bm{g}_{i}^k$
                
                \IF{$k \mod \tau = 0$}
                    \STATE Compute the local estimated Hessian: \newline $\hat{\bm{h}}_{i}^k = \text{Estimator}(\bm{\theta}_i^k)$ using GNB Algorithm \ref{alg:gnb}
                    \STATE Update the local moving average of the \\Hessian estimator:\\ $\bm{h}_{i}^k = \beta_2 \bm{h}_{i}^{k-\tau} + (1 - \beta_2) \hat{\bm{h}}_{i}^k$
                \ELSE
                    \STATE $\bm{h}_{i}^k = \bm{h}_{i}^{k-1}$
                \ENDIF
                
                \STATE Update the local model with weight decay:\newline $\bm{\theta}_i^k = \bm{\theta}_i^k - \eta \lambda \bm{\theta}_i^k$
                \STATE Update the local model:  $\bm{\theta}_i^{k+1} = \bm{\theta}_i^k - \eta \text{clip}\left(\frac{\bm{m}_{i}^k}{\max(\bm{h}_{i}^k, \epsilon)}, \rho\right)$
                \ENDFOR
                \STATE Send the updated local models $\bm{\theta}_i^{k+1}$ back to the server
            \ENDFOR
            
            \STATE Update the global model:  $\bm{\Theta}^{k+1} = \frac{1}{N} \sum_{i=1}^N \bm{\theta}_i^{k+1}$
        \ENDFOR
    \end{algorithmic}
\end{algorithm}

% \subsection{Version 2}

% We introduce another version of federated Sophia. In this version, instead of sharing the model's parameters directly to the server, we share the Hessian and the gradient of each device. This is still communication-efficient since we are only sharing vectors of size d. Sharing the estimated diagonal Hessian along with the gradient allows the server to have all the information needed for more informed decisions in the convergence process. On the server, we propose the following update
% \begin{align}
% \bar{h}^k &= \sum_{i=1}^N h_i^k ,\\
% \bar{m}^k &= \sum_{i=1}^N m_i^k, \\
% \Theta^{k+1} &= \Theta^k - \eta \text{clip}(\frac{\bar{m}^k}{\max(\bar{h}^k, \epsilon)}, \rho),
% \end{align}
% where $\Theta^k$ is the server's model's parameters at communication round $K$.

\subsection{Hessian as a Pre-conditioner}
One problem with first-order methods, like FedAvg, is their inability to incorporate curvature information during the optimization process. Curvature information is best obtained using the Hessian matrix. At the $k$th communication round, a general descent direction can be obtained as follows
\begin{align}\label{direction}
    \Delta\bm{\theta}_i^{k} \triangleq \bm{\theta}_i^{k+1}-\bm{\theta}_i^{k}= (\bm{H}_i^{k})^{-s} \bm{g}_i^k.
\end{align}
Herein, $s = 1$ corresponds to the vanilla Newton method, while $s=0$ rolls back to the gradient descent method. The idea of the Hessian as a pre-conditioner, as shown in \eqref{direction}, is that it adapts the step size taken along the descent direction by $(\bm{H}_i^{k})^{-1}$.

% \subsection*{\textbf{Hessian Estimation}}

% \begin{figure}
%     \centering
%     \includegraphics[width=0.5\linewidth]{Hessian (2).png}
%     \caption{Enter Caption}
%     \label{fig:enter-label}
% \end{figure}

Computing the full Hessian is almost impossible in the case of a very large NN. This makes certain FL algorithms impractical in real-world settings. One way to estimate the Hessian is by computing the diagonal information only\cite{martens2012estimating}. The diagonal elements of the Hessian contain curvature information across each dimension. Thus, \eqref{direction} can be approximated as 
\begin{align}\label{direction_1}
    \Delta\bm{\theta}_i^{k} \approx \textit{diag}(\bm{H}_i^k)^{-1} \bm{g}_i^k.
\end{align}
Several techniques for such diagonal estimation have been proposed, e.g., Hutchinson's and Gauss-Newon-Bartlett (GNB) methods \cite{liu2023sophia}. Fed-Sophia uses the GNB estimator to estimate the Hessian as it has been shown to exhibit a better generalization performance \cite{liu2023sophia}. Given a loss function $f(\bm{\theta})$, the Hessian matrix can be computed using the following Gauss-Newton decomposition
\begin{align}
     \bm{H} = \bm{J}_{\bm{\theta}}\phi(\bm{\theta}, \bm{x}) \cdot \bm{S} \cdot \bm{J}_{\bm{\theta}}\phi(\bm{\theta}, \bm{x})^T + \bm{J}_{\bm{\theta}\bm{\theta}}\phi(\bm{\theta}, \bm{x}) \cdot \bm{q},
\end{align}
where $\bm{J}$ is the Jacobian of $\phi$ with respect to $\bm{\theta}$, $\phi$ is the logits function or the mapping from input to raw output values, $\bm{S}$ is the second-order derivative of the loss with respect to the logits, $\bm{J}_{\bm{\theta}\bm{\theta}}\phi(\bm{\theta}, \bm{x})$ is the second-order derivatives of the logits function with respect to $\bm{\theta}$, and $\bm{q}$ is the first-order derivative of the loss with respect to the logits. The first term is often called the Gauss-Newton matrix, while the second is usually very small and could be negligible. The idea is to estimate the diagonal of the Gauss-Newton matrix and use this as the Hessian's diagonal. The procedure to estimate the diagonal of the Hessian is depicted in Algorithm \ref{alg:gnb}, where $\odot$ denotes the element-wise product. Hence, the estimated Hessian for the $k$th communication round is given by
\begin{align}
    \hat{\bm{h}}_{i}^k = \text{Estimator}(\bm{\theta}_i^k)
\end{align}
where the estimator is the GNB method in our case.

A big advantage of Fed-Sophia is its ability to compute the estimated Hessian's diagonal every $\tau$ iteration, where $\tau$ is an integer usually between 1 and 10. Another advantage for Fed-Sophia is that it uses mini-batches to compute the loss, not the complete data. This might reflect some noise in the gradient and the Hessian. To avoid this issue and smoothen both the gradient and the Hessian, an exponential moving average is computed instead of the raw value for both of them. The update for the gradient is given by
\begin{align}\label{meq}
    \bm{m}_{i}^k = \beta_1 \bm{m}_{i}^{k-1} + (1 - \beta_1) \bm{g}_{i}^k,
\end{align}
while the Hessian is updated, every $\tau$ iterations, using
\begin{align}\label{heq}
    \bm{h}_{i}^k = \beta_2 \bm{h}_{i}^{k-\tau} + (1 - \beta_2) \hat{\bm{h}}_{i}^k,
\end{align}
where $\beta_1$ and $\beta_2$ are two hyperparameters.
\begin{algorithm}[t]
    \caption{Gauss-Newton-Bartlett (GNB)}
    \label{alg:gnb}
    \begin{algorithmic}[1]
        \STATE {\bf Parameters:} $\bm{\theta}$

                \STATE Draw a mini-batch of the input $\{{\bm{x}_b\}}_{b=1}^{B}$
                \STATE Compute the logits on the mini-batch $\{{\phi(\bm{\theta}, \bm{x}_b)\}}_{b=1}^{B}$ 
                \STATE Sample $\hat{\bm{y}}_b \sim \text{Softmax}(\phi(\bm{\theta}, \bm{x}_b))$
                \STATE Calculate $\hat{\bm{g}} = \nabla (1/B \sum f(\phi(\bm{\theta}, \bm{x}_{b}), \hat{\bm{y}_b})$     \STATE return $B \cdot \hat{\bm{g}} \odot \hat{\bm{g}}$
    \end{algorithmic}
\end{algorithm} 
\subsection{Adaptive Step Size}
The update at the $i$th device is performed by dividing the moving average of the gradient by the moving average of the estimated diagonal Hessian, i.e., $\bm{m}_i^k/\bm{h}_i^k$. The motivation for this update is the fact that the ideal step size to take should be proportionally inverse to the Hessian across each dimension to avoid uniform steps across all dimensions. In other words, we need to make weighted steps at each dimension, where the weights are the Hessian's information across those dimensions. Rather than using a fixed update step across all dimensions of the loss function, the idea is to adjust this step to enable sharp dimensions and flat dimensions to behave accordingly. For sharp dimensions, the step size is expected to be less than in flat dimensions. Dividing by the estimated Hessian enables the adaptive step size mechanism, which in turn assigns different step sizes depending on the shape of the loss curvature.

\subsection{Clipping Operation}
Along with the problem of converging to a saddle point in the case of a non-convex loss function, the Hessian entries might be misleading the convergence process in case there are sudden changes along the loss function curvature. To help mitigate these problems, a clipping operation is introduced. The clipping operation guards against inaccurate information that might be incorporated into the Hessian's diagonal. Given a vector $\bm{z}$, the formula for the clipping operation is given by
\begin{align}
    \text{clip}(\bm{z}, \rho) = \max\{\min \{z, \rho\}, -\rho\},
\end{align}
where $\rho$ is a positive real number that controls the maximum update magnitude. Using the clipping operation, the descent direction \eqref{direction_1} becomes
\begin{align}
    \Delta \bm{\theta}_i^k = \text{clip}\left(\frac {\bm{m}_i^k }{\max \{\bm{h}_i^k, \epsilon \}}, \rho\right),
\end{align}
where $\bm{m}_i^k$ and $\bm{h}_i^k$ are computed according to \eqref{meq} and \eqref{heq}, respectively, and $\epsilon$ is a very small positive constant used to prevent division by zero. Algorithm \ref{alg:federated_hess_opt} shows the full details of Fed-Sophia.

\begin{figure*}[t]
\centering
  \begin{subfigure}{0.52\textwidth}
    \centering
    \includegraphics[scale=0.7]{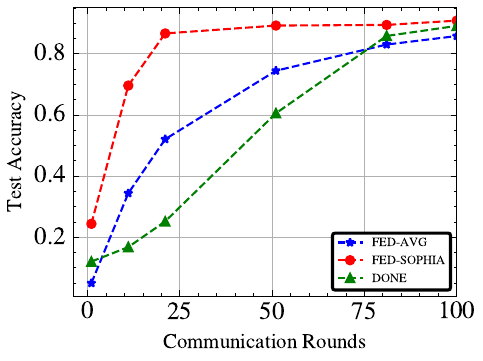}
    \caption{CNN-MNIST} \vspace{0mm}
  \end{subfigure}
  \hspace*{-8em}
  \begin{subfigure}{0.52\textwidth}
    \centering
    \includegraphics[scale=0.7]{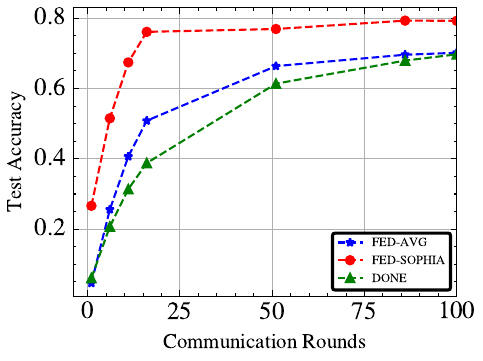}
    \caption{CNN-FMNIST} \vspace{0mm}
  \end{subfigure}
  
  \begin{subfigure}{0.52\textwidth}
    \centering
    \includegraphics[scale=0.7]{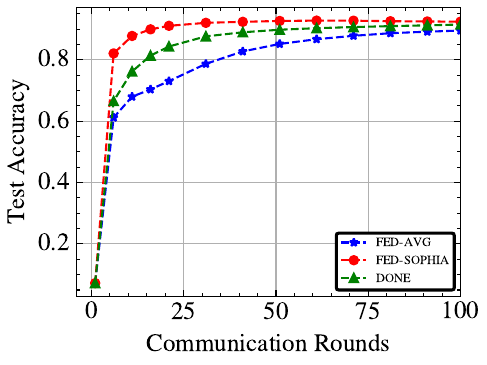}
    \caption{MLP-MNIST} \vspace{0mm}
  \end{subfigure}
  \hspace*{-8em}
  \begin{subfigure}{0.52\textwidth}
    \centering
    \includegraphics[scale=0.7]{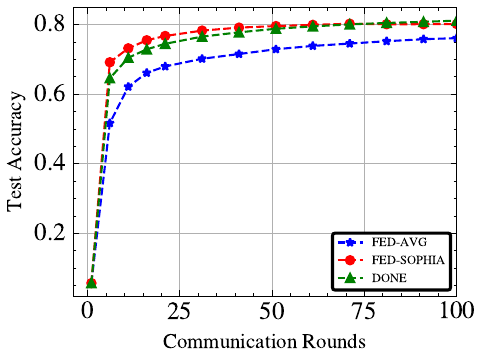}
    \caption{MLP-FMNIST } \vspace{0mm}
  \end{subfigure}
  \vspace{2.5mm}
  \caption{Test accuracy for Fed-Sophia against other baselines in terms of communication rounds for MNIST/FMNIST datasets using MLP and CNN models.}
  \label{fig-2}
\end{figure*}

\section{Numerical Results}\label{RES}
In this section, we lay out the details of the experiments along with the results. First, we compare the performance of Fed-Sophia to the other baselines in terms of communication and computation efficiency. Then, we study the effect of the hyperparameters on Fed-Sophia's performance. Finally, we investigate the energy and carbon footprint of all algorithms.
\subsection{Experimental Settings}
We conducted several experiments to compare the performance of Fed-Sophia with two baselines: FedAvg, a first-order method that utilizes only the gradient information, and DONE, an approximate Newton-type method. Two data sets are used: MNIST and Fashion MNIST. The data is distributed among 32 devices, and each partition is split into $75\%$ and $25\%$ for training and testing, respectively. All the experiments are in the non-IID setting. In our experiments, we consider two different classification models, MLP and CNN, with the loss function being cross-entropy in both of them. The number of local iterations for both Fed-Sophia and FedAvg is taken to be $J=10$. For DONE, we tune the number of local iterations to maintain a fair comparison.

\begin{table}[b]
\caption{Effect of the learning rate and the number of local iterations on the test accuracy for the Fashion MNIST dataset with CNN.}
\begin{center}
\begin{tabular}{|c|c|c|}
\hline

\cline{2-3} 
\textbf{Learning rate ($\eta$)}& \textbf{Local iterations ($J$)}&\textbf{Test accuracy(\%)}\\
\hline
0.01& & 76.3  \\
0.003& 10 & 80.3 \\
0.0005& &  71.9 \\
    
\hline
& 1&58.7  \\

0.001&5 & 73.3 \\
&10 & 76.5 \\
\hline

% \multicolumn{4}{l}%{$^{\mathrm{a}}$Sample of a Table footnote.}
\end{tabular}
\label{tab1}
\end{center}
\end{table}
To monitor the energy footprint of each device $i$, both computation and communication costs are computed as follows
\begin{align}
    E_{total}(k) = E_c(k) + E_t(k),
\end{align}
with  
\begin{align}
    E_c(k) =   \sum_{n=1}^k \sum_{j=1}^J e_{i}^{j,n} \hspace{0.1cm} \text{and} \hspace{0.1cm} E_t(k) =  \sum_{n=1}^k  b(\bm{\theta}_i^{k})e_{i, PS}^k,
\end{align}
where $k$ is the number of communication rounds, $e_{i}^{j,n}$ is the energy consumed by the device $i$ for one local iteration, $b(\bm{\theta}_i^{k})$ is the size of the model vector in bits, and $e_{i, PS}$ is the energy needed to transmit information between the device $i$ and the PS. The transmitted model's parameters are assumed to be in the 32-bit form. We use the same channel model as in \cite{ghalkha2023} and assume a uniform distance between the server and all clients within a space of size $100\times100$ $m^2$. According to Shannon's formula, the maximum achievable rate for each client is given by $R = B \log_2 ( 1 + \frac{P_t}{d_{i, PS} B N_0 }) $, where $B$ is the bandwidth, $P_t$ is the transmission power, $d_{i, PS}$ is the distance between client $i$ and the PS, and $N_0$ is the noise spectral density. We set $P_t = 100$mW, $B=2$MHz, and $N_0=10^{-9}$W/Hz. 

\subsection{Performance Comparison} 
\begin{figure*}[t]
\centering
  \begin{subfigure}{0.52\textwidth}
    \centering
    \includegraphics[scale=0.75]{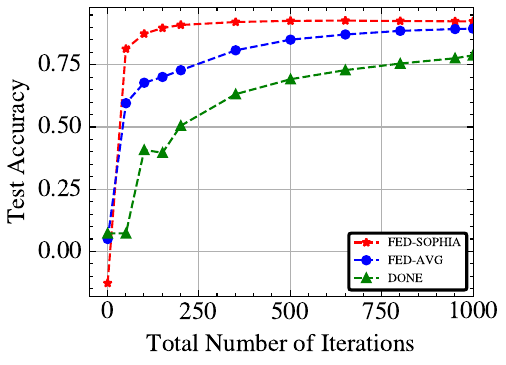}
    \caption{MLP-MNIST}
  \end{subfigure}
  \hspace*{-7em}
  \begin{subfigure}{0.52\textwidth}
    \centering
    \includegraphics[scale=0.75]{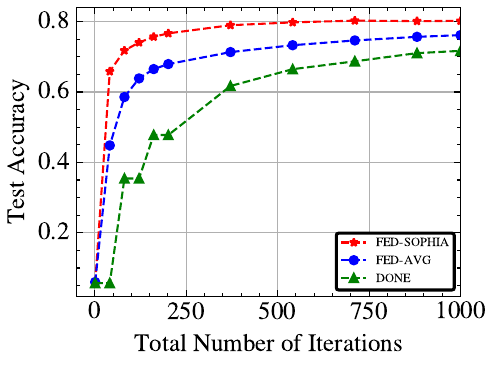}
    \caption{MLP-FMNIST}
  \end{subfigure}
  \caption{Test accuracy for Fed-Sophia against other baselines in terms of the number of total iterations for MNIST and FMNIST datasets using MLP.}
  \label{fig-3}
\end{figure*}

Herein, we compare the performance of Fed-Sophia against FedAvg and DONE in terms of the test accuracy on both MNIST and FMNIST datasets. For Fed-Sophia and FedAvg, we utilize mini-batches of size 512, a common way to train DNNs. On the other hand, DONE requires the full data for each client, which comes at a computational cost. As shown in Fig. \ref{fig-2}, the proposed Fed-Sophia scheme requires fewer communication rounds than FedAvg and DONE in all experiments. It also outperforms both of them in terms of the test accuracy for the CNN model Fig. \ref{fig-2}(a) and 2(b) with the two datasets. For the MLP model, Fed-Sophia outperforms the two baselines with the MNIST dataset, requiring only $30$ communication rounds to converge compared to $70$ and $100$ for DONE and FedAvg, respectively as shown in Fig. \ref{fig-2}(c). For the FMNIST dataset in Fig. \ref{fig-2}(d), DONE and Fed-Sophia are almost identical in terms of test accuracy, with the advantage of early convergence for Fed-Sophia. In Fig. \ref{fig-3}, we plot the test accuracy of all algorithms against the total number of iterations to grasp an idea of the computational cost. For the MNIST dataset, it turns out that Fed-Sophia reaches a target test accuracy of $75\%$ after almost $30$ iterations, while FedAvg and DONE require $200$ and $750$ iterations to achieve the same target accuracy, respectively. On the other hand, Fed-Sophia with FMNIST dataset requires $200$ iterations to converge, while FedAvg and DONE require almost the full number of iterations.
\begin{table*}[t]

\centering

\resizebox{1.9\columnwidth}{!}{\begin{minipage}[t!]{2.0\columnwidth}

\caption{Computation/communication energy costs and corresponding carbon footprints for MNIST with CNN for target accuracy of 75\%.}

\label{Table_1}

\begin{tabularx}{1\linewidth}{p{3.2cm} p{4.2cm} p{4.2cm} p{4.6cm}}

\toprule[1pt]

Algorithm & Computation Energy [MJ] & Communication Energy [MJ] & Total Footprint [kg-CO2-eq] \\

\cmidrule(r){1-1} \cmidrule(r){2-2} \cmidrule(r){3-3} \cmidrule(r){4-4}

{DONE} & $6170E-4$  \hspace{3.5pt}\tikz{

\fill[fill=color3] (0.0,0) rectangle (1.3,0.2);

} & $153.2$ \hspace{2.2pt} \tikz{

\fill[fill=color4] (0.0,0) rectangle (1.1,0.2);

} & $1097E+3$ \\ \hline

{FedAvg} & $2.2E-4$ \hspace{7pt} \tikz{

\fill[fill=color3] (0.0,0) rectangle (0.08,0.2);

} & $139.2$ \hspace{2.5pt} \tikz{

\fill[fill=color4] (0.0,0) rectangle (1.0,0.2);

} & $0.123E+3$ \\ \hline
{Fed-Sophia} & $1.4E-4$ \hspace{7pt} \tikz{

\fill[fill=color3] (0.0,0) rectangle (0.05,0.2);

} & $27.8$ \hspace{6.5pt} \tikz{

\fill[fill=color4] (0.0,0) rectangle (0.2,0.2);

} & $0.004E+3$ \\ \hline

% {ffff} & $7.96E-2$ \hspace{5pt} \tikz{

% \fill[fill=color3] (0.0,0) rectangle (0.56,0.2);

% } & $10.21$ \hspace{5pt} \tikz{

% \fill[fill=color4] (0.0,0) rectangle (0.71
% ,0.2);

% } & $1.83E-4$ \\ 

 & \ \ \ \ \ \ \ \ \ \ \ \ \ \ \,\tikz{

\draw[black] (0.2,0) -- (2.0,0);

\draw[black] (0.2,-2pt) -- (0.2,2pt)node[anchor=north] {\tiny$0$};
\draw[black] (1.0,-2pt) -- (1.0,2pt)node[anchor=north] {\tiny$0.4$};
\draw[black] (1.9,-2pt) -- (1.9,2pt)node[anchor=north] {\tiny$0.8$};

%\draw[black] (2.1,-2pt) -- (2.1,2pt)node[anchor=north] {\tiny$30$};

%\draw[black] (3.6,-0.1pt) -- (3.6,0.1pt)node[anchor=north]{\tiny\ [dBm]};

} \vspace{0mm} & \ \ \ \ \ \ \ \ \,\tikz{

\draw[black] (0.5,0) -- (1.9,0);

\draw[black] (0.5,-2pt) -- (0.5,2pt)node[anchor=north] {\tiny$0$};

\draw[black] (1.2,-2pt) -- (1.2,2pt)node[anchor=north] {\tiny$100$};

\draw[black] (1.9,-2pt) -- (1.9,2pt)node[anchor=north] {\tiny$200$};

% \draw[black] (2.6,-2pt) -- (2.6,2pt)node[anchor=north] {\tiny$5E+20$};

%\draw[black] (3.6,-0.1pt) -- (3.6,0.1pt)node[anchor=north]{\tiny\ [dBm]};

} \vspace{0mm} \\

\bottomrule[1pt]
\label{tab:energy_con}
\end{tabularx}

\end{minipage}}

\vspace{-1cm}
\end{table*}

\subsection{Effect of the Hyperparameters}
% \textcolor{red}{Change caption of Table II.}\\
In this subsection, we study the effect of the learning rate $(\eta)$ and the number of local iterations $(J)$ on the performance of Fed-Sophia. We fix $J=10$ and experiment with different learning rates for the Fashion MNIST dataset. The test accuracy for the three different values of the learning rate is summarized in Table \ref{tab1}. Next, we study the effect of the local iterations to see if it is worth adding more computational overhead to achieve some performance gain. As shown in Table \ref{tab1}, as we increase the local number of iterations, Fed-Sophia gains a reasonable performance. It is also clear that the performance gain between 5 and 10 local iterations is minimal.
%, hence a small number of local iterations is all that Fed-Sophia needs.

%--------Table packages

\usetikzlibrary{patterns,arrows}
\definecolor{color1}{RGB}{237, 191, 193}
\definecolor{color2}{RGB}{229, 153, 157}
\definecolor{color3}{RGB}{225, 123, 116}
\definecolor{color4}{RGB}{10,10,200}
\definecolor{color5}{RGB}{203,52,38}

% \begin{figure}
%     \centering
%     \includegraphics[width=0.5\linewidth]{total_iter.png}
%     \caption{Enter Caption}
%     \label{fig:enter-label}
% \end{figure}

\subsection{Energy-Efficiency and Carbon Footprint}
As shown in Table \ref{Table_1}, Fed-Sophia outperforms DONE and FedAvg in terms of computation energy consumption for a target test accuracy of 75\% using a CNN model for the MNIST dataset. Furthermore, DONE consumes up to $4000\times$ computational power than Fed-Sophia. Fed-Sophia requires almost half of the computational energy needed for FedAvg. On the other hand, the communication energy for Fed-Sophia is less than the other two baselines by a large factor. Fed-Sophia only requires almost 20\% of the needed communication energy for the closest baseline, i.e., FedAvg. Therefore, the total footprint for Fed-Sophia is minimal, which proves the communication and computation efficiency of our proposed method compared to the other two baselines.

\section{Conclusion}
In this paper, we presented Fed-Sophia, a scalable and second-order method for FL. Fed-Sophia is a computation and communication-efficient method that leverages the curvature information of the loss function by estimating the diagonal of the Hessian. The update is the moving average of the gradient scaled by the inverse of the estimated diagonal Hessian. Furthermore, Fed-Sophia mitigates the effect of inaccurate entries of the Hessian by augmenting the update into a clipping operation, guarding against misleading curvature information. Our experiments showed that the proposed method outperforms other baselines on image classification tasks. %Employing Fed-Sophia to train an LLM in a federated manner and convergence analysis of Fed-Sophia are two potential research directions for the future.

\section*{Acknowledgments} \vspace{0mm}
This work is partially supported by the European Commission through Grant no. 101095363 (Horizon Europe SNS JU ADROIT6G project), Grant no. 101139266 (6G-INTENSE project), and the Academy of Finland, 6G Flagship program (Grant no. 346208).

\bibliographystyle{IEEEtran}
\bibliography{references} 

\end{document}